\documentclass[10pt,twocolumn,letterpaper]{article}

\usepackage{cvpr}
\usepackage{times}
\usepackage{epsfig}
\usepackage{graphicx}
\usepackage{amsmath}
\usepackage{amssymb}
\usepackage{pifont}
\usepackage{tikz} 
\usepackage{xcolor}
\usepackage{pgfplots}
\usepackage{subcaption} 
\usepackage{pgfplotstable}
\usepackage{comment}
\usepackage{url}
\usepackage{ctable} 
\usepackage{lipsum}

\usepackage[pagebackref=true,breaklinks=true,letterpaper=true,colorlinks,bookmarks=false]{hyperref}

\cvprfinalcopy 

\def\cvprPaperID{3915} 

\pgfplotstableread[col sep=comma]{tables/mAP_all.csv}\datatable
\makeatletter
\pgfplotsset{
    /pgfplots/flexible xticklabels from table/.code n args={3}{%
        \pgfplotstableread[#3]{#1}\coordinate@table
        \pgfplotstablegetcolumn{#2}\of{\coordinate@table}\to\pgfplots@xticklabels
        \let\pgfplots@xticklabel=\pgfplots@user@ticklabel@list@x
    }
}

\pgfplotstableread[col sep=comma]{tables/oth_all.csv}\datatableoth
\makeatletter
\pgfplotsset{
    /pgfplots/flexible xticklabels from table/.code n args={3}{%
        \pgfplotstableread[#3]{#1}\coordinate@table
        \pgfplotstablegetcolumn{#2}\of{\coordinate@table}\to\pgfplots@xticklabels
        \let\pgfplots@xticklabel=\pgfplots@user@ticklabel@list@x
    }
}

\begin{document}

\title{RetinaMask: Learning to predict masks improves state-of-the-art single-shot detection for free}

\author{Cheng-Yang Fu \quad Mykhailo Shvets 
\quad Alexander C. Berg\\
Computer Science Department of \\
UNC at Chapel Hill\\
{\tt\small \{cyfu, mshvets,  aberg\}@cs.unc.edu}}



\maketitle

\begin{abstract}
 Recently two-stage detectors have surged ahead of single-shot detectors in the accuracy-vs-speed trade-off.  Nevertheless single-shot detectors are immensely popular in embedded vision applications.  This paper brings single-shot detectors up to the same level as current two-stage techniques.   We do this by improving training for the state-of-the-art single-shot detector, RetinaNet, in three ways: integrating instance mask prediction for the first time, making the loss function adaptive and more stable, and including additional hard examples in training.  We call the resulting augmented network RetinaMask.  The detection component of RetinaMask has the same computational cost as the original RetinaNet, but is more accurate.  COCO test-dev results are up to 41.4 mAP for RetinaMask-101 vs 39.1mAP for RetinaNet-101, while the runtime is the same during evaluation. Adding Group Normalization increases the performance of RetinaMask-101 to 41.7 mAP. Code is at: \url{https://github.com/chengyangfu/retinamask}
\end{abstract}

\vspace{-.1cm}
\section{Introduction}
\vspace{-.1cm}

Single-shot detectors~\cite{redmon2015yolo,redmon2017yolov2,liu2016ssd,fu2017dssd,lin2017focal} have become extremely popular in applications where speed and computational resources are important design considerations. These include embedded vision applications, self-driving cars, and mobile phone vision.  Despite this intense usage, there has been little improvement in state-of-the-art performance for single-shot detectors, e.g. RetinaNet~\cite{lin2017focal}.  When published in 2017, RetinaNet effectively cleaned up a range of work on single-shot detection, building on~\cite{redmon2017yolov2} and~\cite{liu2016ssd} and introducing innovations in training that resulted in state-of-the-art performance in terms of the speed-versus-accuracy trade-off (sharing the frontier with YOLOv3~\cite{redmon2018yolov3} on the faster, lower-accuracy, end of the spectrum, and Mask R-CNN on the high end).  While there have not been significant improvements in performance on top of RetinaNet, two-stage detectors have advanced over the intervening time, and now outperform RetinaNet on the speed-vs-accuracy trade-off.  Part of this improvement has been due to architectures like Mask R-CNN that allow training multiple prediction heads on top of the region proposal and bounding box prediction stages of the detector~\cite{he2017maskrcnn,guler2018densepose}.

\begin{figure}
\resizebox {\columnwidth} {!} {
\begin{tikzpicture}[y=.8cm, x=.03cm,font=\sffamily]
    \draw[black!30, dashed] (0, 1) -- (140, 1);
    \draw[black!30,dashed] (0, 2) -- (140, 2);
    \draw[black!30,dashed] (0, 3) -- (140, 3);
    \draw[black!30,dashed] (0, 4) -- (250, 4);
    \draw[black!30,dashed] (0, 5) -- (250, 5);
    \draw[black!30,dashed] (0, 6) -- (250, 6);

	\draw (0,0) -- coordinate (x axis mid) (250,0);
    	\draw (0,0) -- coordinate (y axis mid) (0,7);
    	\foreach \x in {50,100,150,200,250}
     		\draw (\x,1pt) -- (\x,-3pt)
			node[anchor=north] {\x};
    	\foreach \y\ytext in {0/28, 1/30, 2/32, 3/34, 4/36, 5/38, 6/40}
     		\draw (1pt,\y) -- (-3pt,\y) 
     			node[anchor=east] {\ytext}; 
     
	\node[below=0.8cm] at (x axis mid) {Inference time [ms]};
	\node[rotate=90, above=0.8cm] at (y axis mid) {Accuracy [mAP]};

	plots
	\draw[blue!30, thick] plot[mark=*, mark options={fill=blue!30}] 
		file {RetinaNet-50};
	\draw[red!30, thick] plot[red, mark=*, mark options={fill=red!30} ] 
		file {RetinaNet101};
	\draw[orange, thick] plot[mark=*, mark options={fill=orange} ] 
		file {yolov3};
	\draw[blue, thick] plot[mark=square*, mark options={fill=blue}] 
		file {sup-RetinaMask-50};
	\draw[red, thick] plot[mark=square*, mark options={fill=red}] 
		file {sup-RetinaMask-101};
	\draw plot[mark=*, mark options={fill=green}]
		file {MaskRCNN-101-paper};
	\draw plot[mark=*, mark options={fill=green}]
		file {MaskRCNN-101-detectron};
	\draw plot[mark=*, mark options={fill=green}]
		file {MaskRCNN-101-pytorch};
    
	legend
	\begin{scope}[shift={(150,0.5)}] 
	\draw (0,0) -- 
		plot[mark=*, mark options={fill=blue!30}] (0.25,0) -- (0.5,0) 
		node[right]{\cite{lin2017focal}RetinaNet-50};
	\draw[yshift=\baselineskip] (0,0) -- 
		plot[mark=*, mark options={fill=red!30}] (0.25,0) -- (0.5,0)
		node[right]{\cite{lin2017focal}RetinaNet-101};
	\draw[yshift=2\baselineskip] (0,0) -- 
		plot[mark=*, mark options={fill=orange}] (0.25,0) -- (0.5,0)
		node[right]{\cite{redmon2018yolov3}YOLOv3};
	\draw[yshift=3\baselineskip] (0,0) -- 
		plot[mark=*, mark options={fill=green}] (0.25,0) -- (0.5,0)
		node[right]{\cite{he2017maskrcnn}Mask R-CNN-101};
	\draw[yshift=4\baselineskip] (0,0) -- 
		plot[mark=square*, mark options={fill=blue}] (0.25,0) -- (0.5,0)
		node[right]{RetinaMask-50};
	\draw[yshift=5\baselineskip] (0,0) -- 
		plot[mark=square*, mark options={fill=red}] (0.25,0) -- (0.5,0)
		node[right]{RetinaMask-101};
	\end{scope}
\end{tikzpicture}
}
\caption{Accuracy versus inference time on COCO \texttt{test-dev}. For fair comparison with RetinaNet, we train our RetinaMask models at \{400, 500, 600, 700, 800\} resolution without using multi-scale augmentation during training. Our best model, which achieves 42.6~mAP, can be found in Table~\ref{table:compare_stoa}. Our improved versions, RetinaMask-50/101 respectively, are shown with blue/red square markers.  The original RetinaNet-50/101 results are shown with blue/red circle markers. The state-of-the-art two-stage detector Mask R-CNN has improved since publication. We show three versions with green circle markers: original paper 38.5/195 detection+mask/M40 (mAP/ms/GPU), Detectron  \cite{maskrcnndetectronurl} Caffe2  (40.0/119 detection only/P100), and PyTorch \cite{maskrcnnpytorchurl} (40.1/143 detection only/V100).}
\label{fig:speed_vs_accuracy}
\vspace{-2.5mm}
\end{figure}
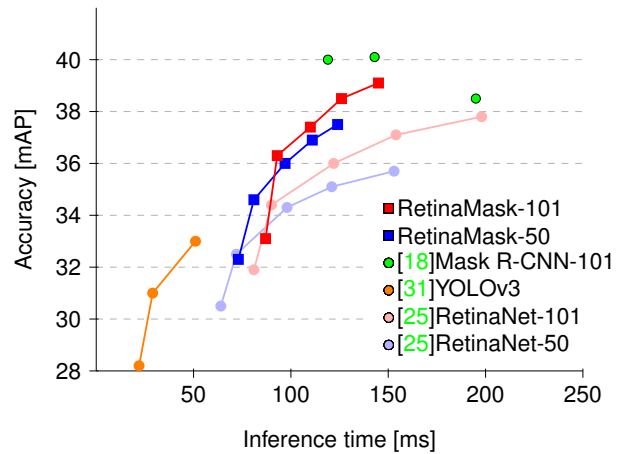


In this paper we show how to improve the accuracy of state-of-the-art single-shot detectors (e.g. RetinaNet~\cite{lin2017focal}, SSD~\cite{liu2016ssd,fu2017dssd}) largely by adding the task of {\bf instance mask prediction during training}, but also by introducing an {\bf adaptive loss} that improves robustness to parameter choice during training, and including {\bf more difficult examples} in training.  We call the resulting system RetinaMask to make clear that its training included the additional task of instance mask prediction, which in the past had been added to two-stage but not to single-shot detectors.  These modifications involve more work during training, but it is possible to evaluate just the detection part of RetinaMask, which has exactly the same computational cost as the original RetinaNet detector---our modifications to training result in detectors with better accuracy at the same computational cost.

Because we keep the structure of the detector at test time unchanged, our approach can be directly applied to the wide range of embedded applications that choose single-shot detectors over two-stage detectors.  This is a subtle point, but today (before this paper), two-stage detectors significantly beat single stage detectors on the speed-vs-accuracy trade-off on a standard desktop/workstation + high-end GPU configurations.  However, when adapting these implementations to lower-power embedded devices the cost of resampling (e.g. ROI-Align) for the second stage is often exacerbated by communications between
the hardware that is used to accelerate convolutions and the CPU. Single-shot approaches avoid this extra implementation challenge, which makes them popular in many implementations.

This allusion to implementation overhead for a variety of special purpose architectures brings up one of the difficulties of keeping track of progress on the speed-vs-accuracy trade-off in detection.  In addition to the algorithms, the hardware itself and software libraries are constantly evolving year over year.  These complexities are small relative to the variety of software and architectures for embedded systems.  The main point of our work is to show a significant improvement in the accuracy of single-shot detectors with keeping the same computational cost as the original network during inference. Note the the plot in Figure~\ref{fig:speed_vs_accuracy} shows an improvement in speed as well, but this is partly due to the next generation of libraries and slightly different hardware.  We expect the improvements demonstrated in this paper to be applicable to a wide range of single-shot detector implementations.

In summary, this paper shows a significant increase in the accuracy of state-of-the-art single-shot detectors by introducing three new techniques that improve training:
\vspace{-.1cm}
\begin{enumerate}
\vspace{-.1cm}
    \item Adding a novel instance mask prediction head to the single-shot RetinaNet detector during training.
 \vspace{-.1cm}
   \item A new self-adjusting loss function that improves robustness during training.
\vspace{-.1cm}
    \item Including more of the positive examples in training, even those with low overlap.
\end{enumerate}
\vspace{-.1cm}
Each of these contributions is analyzed with ablation studies, and together they provide a large boost to the accuracy of RetinaNet, bringing it up to state-of-the-art accuracy again.

\vspace{-.1cm}
\section{Related Work}
\vspace{-.1cm}

We review recent developments in object detection using two-stage and single-shot techniques, general techniques for improving detection, and work on integrating instance mask prediction with detection.

\smallskip
\noindent {\bf Two-Stage Detectors:}
Two-stage detectors follow a long line of reasoning in computer vision about grouping and perception. They first propose potential object locations in an image---region proposals---and then apply a classifier to these regions to score potential detections.  Earlier sliding-window approaches ran into scaling problems as the number of potential windows combined with the number of models became unmanageable~\cite{dalal_hog,felzenszwalb2010dpm}.  Selective Search~\cite{Uijlings2013selectivesearch} allowed more expensive and accurate bag-of-visual-words (BoVW) features to be considered by using low-level vision to identify a smaller number of potential locations that needed to be evaluated. A transition to deep learning models was done with R-CNN~\cite{girshick2014rcnn}, which used a convolutional neural network to replace the BoVW, resulting in a significant accuracy improvement. SPPNet~\cite{he2015sppnet} sped up this process by direct region pooling on the feature layers instead of repetitive image cropping. Then Fast R-CNN~\cite{girshick15fastrcnn}, and Faster R-CNN~\cite{ren2015faster} sped up detection even further and increased accuracy by replacing Selective Search with a Regional Proposal Network. Faster R-CNN was also the first end-to-end trainable deep learning model for object detection. R-FCN~\cite{dai2016rfcn} used position-sensitive features and ROI-pooling to make a fully convolutional network design.


\smallskip
\noindent {\bf Single-Shot Detectors:}
In contrast to two-stage detectors,  single-shot detectors avoid image or feature re-sampling. OverFeat~\cite{sermannet2014overfeat} and DeepMultiBox~\cite{erhan2014deepmultibox} were early examples. Then, YOLO~\cite{redmon2015yolo, redmon2017yolov2} and SSD~\cite{liu2016ssd} popularized the single-shot approach by demonstrating models that ran in real-time with good accuracy. More recently, RetinaNet~\cite{lin2017focal} proposed Focal Loss to address the extreme class imbalance problem between target and background, and cleaned up a number of design aspects for single-shot detection.

\smallskip
\noindent {\bf Techniques for Improving Detectors:}
Several techniques for improving detection apply to both Single-Shot and Two-Stage Detectors.  First, cleaner training data often helps achieve faster convergence and higher final accuracy. Online Hard Negative Sampling~\cite{shrivastave2016ohem} uses non-maximum suppression (nms) during training to provide negative examples diversity. Second, certain model modifications add context information to predictions. SSD~\cite{liu2016ssd} and MS-CNN~\cite{cai16mscnn} both predict instances across features layers of different resolutions. DSSD~\cite{fu2017dssd}, Feature Pyramid Network~\cite{lin2016fpn} and TDN~\cite{shrivastava2016tdm} combine feature layers in a top-down manner to enrich the context of coarser features, strengthening the feature representation for better detection. Also, additional training information is beneficial for detectors. BlitzNet~\cite{dvornik17blitznet} augments SSD with a semantic segmentation prediction branch, combining these two tasks in a single network, which results in higher detection accuracy.

\smallskip
\noindent {\bf Instance Segmentation:}
As object detection matured and demonstrated strong accuracy and high speed, the research community started shifting attention to the more detailed task of instance segmentation. In addition to generating a tight bounding box for each object, instance segmentation requires a pixel-level mask for that object.
The COCO~\cite{lin2014coco} dataset established a recognized benchmark for this task by holding the Instance Segmentation Challenge starting in 2015. Current state-of-the-art instance segmentation approaches are based on two-stage detectors, adding an instance mask prediction module after detection. MNC~\cite{dai2016instance} breaks down the Instance Segmentation into three stages, namely object detection, class-agnostic mask prediction, and mask categorization. FCIS~\cite{li2016fcis} extends the idea of  R-FCN~\cite{dai2016rfcn} by using position-sensitive score maps for mask prediction. The recent Mask R-CNN~\cite{he2017maskrcnn} identifies the core issue for mask prediction in ROI pooling box misalignment, which arises from pooling box quantization over the coarse feature scale. Bilinear interpolation is introduced in their ROI-Align module to fix this issue. 
Mask R-CNN has been further improved in the Path Aggregation Network~\cite{liu2018pathnet}, using the ROI-Align operation on multiple feature layers to aggregate better features for instance segmentation.


\vspace{-.1cm}
\section{Model}
\label{sec:model}
\vspace{-.1cm}

We start with the RetinaNet settings in Detectron\footnote{\url{https://github.com/facebookresearch/Detectron}} and rebuild it in PyTorch to form our baseline. Then, we introduce the following modifications to the baseline settings: best matching policy (Sec.~\ref{sec:best_matching}), and modified bounding box regression loss (Sec.~\ref{sec:selfadjust}). Finally, we describe how to add the mask prediction module on top of RetinaNet (Sec.~\ref{sec:maskpredict}).

\vspace{-.1cm}
\subsection{Best Matching Policy}
\label{sec:best_matching}
\vspace{-.1cm}

In the bounding box matching stage, the RetinaNet policy is as follows. All anchor boxes that have an intersection-over-union (IOU) overlap with a ground truth object greater than 0.5, are considered positive examples. If the overlap is less than 0.4, the anchor boxes are assigned a negative label. All anchors for which the overlap falls between 0.4 and 0.5 are not used in the training.
However, there exists an exceptional case for which the assignment can be improved. Specifically, some of the ground truth objects' aspect ratios are outliers, with one side much longer than the other. Thus, no anchor box can be matched to those according to the RetinaNet strategy. For each of these ground truth boxes we propose to find its best matching anchor box, relaxing the overlapping IOU threshold. We provide an ablation study using different thresholds on the best matching anchors. The results suggests that using best matching anchor with any nonzero overlap gives the best accuracy (notice that such anchors always exists, because single-shot anchors are densely sampled).

\begin{figure}[h]
\centering
\begin{subfigure}[b]{0.4\linewidth}
\begin{tikzpicture}
      \draw[->] (-0.2,0) -- (2.2,0) node[right] {$x$};
      \draw[->] (0,-0.2) -- (0,2.2) node[above] {$y$};
      \draw[scale=1.0,domain=0:1,smooth,variable=\x,blue] plot ({\x},{\x*\x*0.5});
      \draw[scale=1.0,domain=1:2.2,smooth,variable=\x,blue] plot ({\x},{-0.5+\x});
      \node at (1.0,0.5) [circle,fill,inner sep=1.5pt]{};
      \draw[dotted] (1,-0.2) -- (1,2.2) node[above] {$\beta$};
      
      \draw[dashed, <->] (0.01,1.75) -- (1.0, 1.75) ;
      \draw[dashed, <->] (1.0, 1.75) -- (2.2, 1.75);
     \node at (0.5, 2.0)  {L2}; 
     \node at (1.5, 2.0)  {L1}; 
      
\end{tikzpicture}
\caption{\footnotesize Smooth L1}
\label{fig:smoothl1}
\end{subfigure}
\hspace{1em}
\begin{subfigure}[b]{0.45\linewidth}
\begin{tikzpicture}
      \draw[->] (-0.2,0) -- (2.2,0) node[right] {$x$};
      \draw[->] (0,-0.2) -- (0,2.2) node[above] {$y$};
      \draw[scale=1.0,domain=0:1,smooth,variable=\x,blue] plot ({\x},{\x*\x*0.5});
      \draw[scale=1.0,domain=1:1.5,smooth,variable=\x,blue] plot ({\x},{-0.5+\x});
      \node at (1.0,0.5) [circle,fill,inner sep=1.5pt]{};
      \draw[dotted] (1,-0.2) -- (1,2.2) node[above] {$\beta$};
      \draw[dotted] (1.5, -0.2) -- (1.5,2.2) node[above] {$\hat{\beta}$};
      \draw[dashed, <->] (0.01,1.75) -- (0.98, 1.75);
      \draw[dashed, <->] (1.02,1.75) -- (1.5, 1.75);
      \draw[dashed, ->] (1.0, 0.5) -- (0.4, 0.5) node[above] {\scriptsize Moving};
     \node at (0.5, 2.0)  {L2}; 
     \node at (1.3, 2.0)  {L1}; 
\end{tikzpicture}
\caption{\footnotesize Self-Adjusting Smooth~L1}
\label{fig:adjust_smoothl1}
\end{subfigure}

\caption{Smooth L1 and Self-Adjusting Smooth~L1. In Smooth~L1 Loss (a) $\beta$ if a fixed threshold that separates loss into L2 and L1 regions. In the proposed Self-Adjusting Smooth~L1 (b), the $\beta$ is calculated as the difference between running mean and running variance of L1~loss and the value is clamped to the [0, $\hat{\beta}$] range. The $\beta$ is approaching 0 during training.}
\vspace{-2.5mm}
\end{figure}
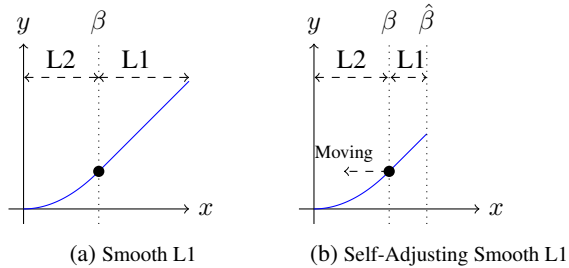

\vspace{-.1cm}
\subsection{Self-Adjusting Smooth~L1 Loss}
\label{sec:selfadjust}
\vspace{-.1cm}

Smooth~L1 Loss for object detection was originally proposed in Fast R-CNN~\cite{girshick15fastrcnn} to make bounding box regression more robust by replacing the excessively strict L2 Loss. Current state-of-the-art methods such as Faster R-CNN~\cite{ren2015faster}, R-FCN~\cite{dai2016rfcn} and SSD~\cite{liu2016ssd} still use this loss.

In Smooth~L1 Loss described in Equation~\ref{eq:smoothl1}, a point $\beta$ splits the positive axis range into two parts: $L2$ loss is used for targets in range $[0, \beta]$, and $L1$ loss is used beyond $\beta$ to avoid over-penalizing  outliers. The overall function is smooth (continuous, together with its derivative), as illustrated in Figure~\ref{eq:smoothl1}. However, the choice of control point($\beta$) is heuristic and is usually done by hyper parameter search.
\vspace{-.1cm}
\begin{equation}
f(x) = \begin{cases}
0.5  \frac{x^{2}}{\beta}, & \text{if} \ |x| < \beta \\
|x| -0.5\beta, & \text{otherwise}
\end{cases}
\label{eq:smoothl1}
\end{equation}
We propose an improved version of Smooth~L1 called Self-Adjusting Smooth~L1 Loss. Inside the loss function, the running mean and variance of the absolute loss are recorded. We use the running minibatch mean and variance with \texttt{momentum}=0.9 to update these two parameters.

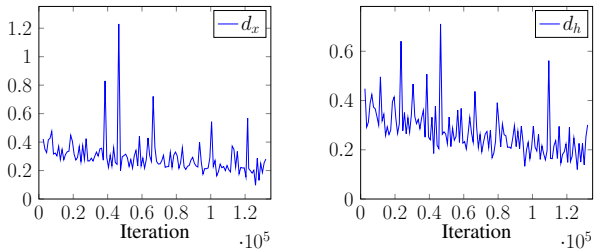
\begin{figure}[h]
\centering
\begin{subfigure}[b]{0.45\linewidth}
\begin{tikzpicture}[scale=0.45]
\LARGE
\begin{axis}
[xmin=0, ymin=0, xmax=135000, xlabel={Iteration}]
\addplot[blue, thick]  table [x=i, y=m0, col sep=comma]  {tables/smoothl1_running.csv};
\legend{$d_x$};
\end{axis}
\end{tikzpicture}
\end{subfigure}
\hspace{1em}
\begin{subfigure}[b]{0.45\linewidth}
\begin{tikzpicture}[scale=0.45]
\LARGE
\begin{axis}
[xmin=0, ymin=0, xmax=135000, xlabel={Iteration}]
\addplot[blue, thick]  table [x=i, y=m2, col sep=comma]  {tables/smoothl1_running.csv};
\legend{$d_h$};
\end{axis}
\end{tikzpicture}
\end{subfigure}
\caption{Running mean of the L1~loss applied to bounding box regression variables: $x$ offset prediction d$_x$ and width prediction d$_w$. Similar plots for $y$ offset prediction d$_y$ and height prediction d$_h$ are omitted for readability.}
\label{fig:running_mean}
\vspace{-2.5mm}
\end{figure}

Then, the parameters are used to calculate the control point. Specifically, the control point is chosen to be equal to the difference between the running mean and running variance ($\mu_{R}- \sigma_{R}^2$), and the value is clipped to a range $[0, \hat{\beta}]$, as can be seen in Equation~\ref{eq:self_beta}. Clipping is used because running mean is unstable during training, as the number of positive examples in each batch is different. Figure~\ref{fig:running_mean} shows the running mean of L1 loss for the x offset and for width adjustment prediction in bounding box regression.  We observe a decreasing trend for both during training.

\vspace{-.1cm}
\begin{equation}
\begin{split}
  \mu_B &= \frac{1}{n} \sum_{i=1}^{n}|x_i|, \quad \sigma_B^2 = \frac{1}{n} \sum_{i=1}^{n}(|x_i| - \mu_B)^2 \\
  \mu_R &=  \mu_R * \text{m} + \mu_B*(1-\text{m})\\
  \sigma_R^2 &= \sigma_R^2 * \text{m} + \sigma_B^2 * (1-\text{m}) \\
  \beta &= \text{max}(0, \text{min}(\hat{\beta}, \mu_R - \sigma_R^2))
\end{split}
\label{eq:self_beta}
\end{equation}
\vspace{-.1cm}

\vspace{-.1cm}
\subsection{Mask Prediction Module}
\label{sec:maskpredict}
\vspace{-.1cm}

\begin{figure*}
\includegraphics[width=\textwidth]{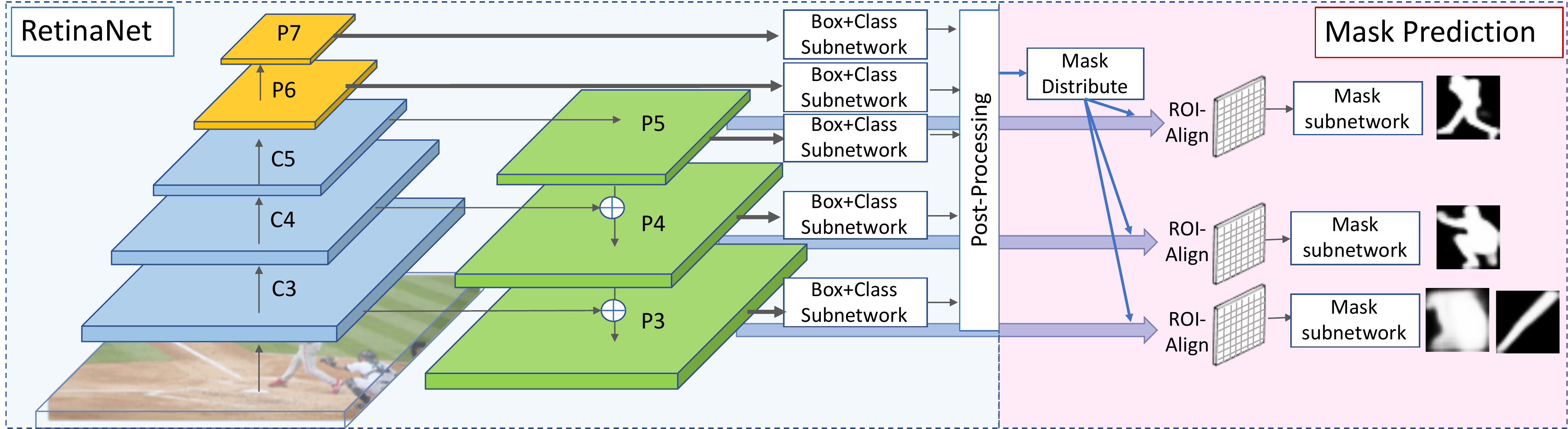}

\caption{RetinaMask network architecture. This figure demonstrates a single-shot detector extended with the mask prediction module. Left part shows the RetinaNet on the Feature Pyramid Network.  The classification and bounding box regression module is applied on the feature layers, \{$P_3$, $P_4$, $P_5$, $P_6$, $P_7$\}. After the predicted bounding boxes are processed and aggregated, they are distributed to \{$P_3$, $P_4$, $P_5$\} for mask predictions according to the size. This results in $P_5$ predicting masks for larger objects, and $P_3$ predicting smaller objects.}
\label{fig:arch}
\vspace{-2.5mm}
\end{figure*}

In order to add the mask prediction module, single-shot detection predictions are treated as mask proposals. After running RetinaNet for bounding box predictions, we extract the top $N$ scored predictions. Then, we distribute these mask proposals to sample features from the appropriate layers of the FPN according to Equation~\ref{eq:mask_distribute} proposed in FPN~\cite{lin2016fpn}. Figure~\ref{fig:arch} illustrates the assignment process.   We use the following equation to determine which feature map, $P_k$ to sample from for predicting the instance mask:

\vspace{-.1cm}
\begin{equation} 
k = {\lfloor}k_0 + \log_{2}{\sqrt{wh}/224}{\rfloor}, \\
\label{eq:mask_distribute}
\end{equation}
where k$_{0}=4$, and w, h are the width and height of the detection. If the size is smaller than $224^2$, it will be assigned to feature layer $P_3$, between $224^2$ to $448^2$ is assigned to $P_4$, and larger than $448^2$ is assigned to $P_5$. 

In our final model, we use the \{$P_3$, $P_4$, $P_5$, $P_6$, $P_7$ \} \footnote{We use the same definition as in FPN~\cite{lin2016fpn} and RetinaNet~\cite{lin2017focal}.} feature layers for bounding box predictions and \{$P_3$, $P_4$, $P_5$\} feature layers for mask prediction. In our ablation study, we analyze the impact of using more feature layers for mask proposals assignment, showing that this does not give any performance boost.


\smallskip
\noindent {\bf Network:}
Figure~\ref{fig:arch} shows a high-level overview of the model. We use ResNet-50 and ResNet-101 as backbone models in our experiments, freezing all of the Batch Normalization layers. Following the Feature Pyramid Network~\cite{lin2016fpn} setting, we add extra layers ($P_6$ and $P_7$) and form top-down connections ($P_5$, $P_4$, and $P_3$). The dimensionality of each of the Feature Pyramid layers ($P_{3}, \dots, P_{7}$) is set to 256.
The bounding box classification head consists of 4 convolutional layers (conv3x3(256) + ReLU) and uses 1 convolution (conv3x3(number of anchors * number of classes)) with point-wise sigmoid nonlinearities. 
For bounding box regression,  we adopt the class-agnostic setting.  We also run 4 convolutional layers (conv3x3(256) + ReLU) and 1 output layer (conv3x3(number of anchors * 4)) to refine the anchors.
Once the bounding boxes are predicted, we aggregate them and distribute to the Feature Pyramid layers, as discussed above. The ROI-Align operation is performed at the assigned feature layers, yielding 14x14 resolution features, which are fed into 4 consequent convolutional layers (conv3x3), and a single transposed convolutional layer (convtranspose2d 2x2) that upsamples the map to 28x28 resolution. Finally, a prediction convolutional layer (conv1x1) is applied. We predict class-specific masks.

\vspace{-.1cm}
\subsection{Training}
\vspace{-.1cm}

To train RetinaNet, we follow the settings in the original paper. 
Images are resized to make the shorter side equal to 800 pixels, while limiting the longer side to 1333 pixels.

We use batch size of 16 images, weight decay $10^{-4}$, momentum $0.9$, and train for 90k iterations with the base learning rate of $0.01$,  dropped to $0.001$ and $0.0001$ at iterations 60k and 80k. We train our models on servers with 4 1080Ti GPUs. For some models (e.g. a ResNet-101-FPN backbone), there is not enough GPU memory for this batch size. If this is the case, we follow the Linear Scaling policy proposed in~\cite{goyal2017imagenet1hr} and reduce the batch size (increasing the number of training iterations and reducing the learning rate accordingly). In order to train with the Mask Prediction Module, we extend the number of training iterations by a factor of 1.5x, or 2x, during multi-scale training. The multi-scale training is done at scales \{640, 800, 1200\}.

Thus, for 1.5x training, the number of iterations is set to 135k, and learning rate drops occur at iterations 90k, and 120k.  For 2x, we train the network 180k iterations, and drop the learning rate at 120k, 160k. The training time is $\approx$ 56 hours when using ResNet-50 with 1.5x train iteration, while for ResNet-101 it goes up to $\approx$ 75 hours, for the same number of iterations.

The  anchor boxes span 5 scales and 9 combinations (3 aspect ratios [0.5, 1, 2] and 3 sizes [$2^0$, $2^{1/3}$ , $2^{2/3}$ ]), following ~\cite{lin2017focal}. The base anchor sizes range from $32^2$ to $512^2$ on Feature Pyramid levels $P_3$ to $P_7$ . Each anchor box is matched to no more than one ground truth bounding box.
The anchors that have intersection-over-union overlap with a ground truth box larger than 0.5 are considered positive examples. On the other hand, if the overlap is less than 0.4, such anchors are treated as negative examples. Then, we use the proposed best matching policy, as described in Section~\ref{sec:model}, which can only add positive examples.
For the Focal Loss function~\ref{eq:focalloss} used in classification, we set $\alpha=0.25$, $\gamma=2.0$, and initialize the prediction logits according to $\mathcal{N}(0, 0.01)$ distribution. For the bounding box regression we add the proposed Self-Adjusting Smooth L1 and limit the control point to the range $[0, 0.11]$ ($\hat{\beta}=0.11$). 
\vspace{-.1cm}
\begin{equation} 
\texttt{FL} = -\alpha_{t}(1 - p_{t})^{\gamma}\log(p_t) \\
\label{eq:focalloss}
\end{equation}
For each image during training, we also run suppression and top-100 selection of the predicted boxes (the same processing as single-shot detectors apply during inference). Then, we add ground truth boxes to the proposals set, and run the mask prediction module. Thus, the number of mask proposals is (100+Gt) during training. The final loss function is a sum of the three losses: \texttt{$Loss_{boxCls}$} + \texttt{$Loss_{boxReg}$} + \texttt{$Loss_{mask}$}.

\vspace{-.1cm}
\subsection{Inference}
\vspace{-.1cm}
First, during the bounding box inference we use a confidence threshold of $0.05$ to filter out predictions with low confidence. Second, we select the top 1000 scoring boxes from each prediction layer. Third, we apply non-maximum suppression~(nms) with threshold $0.4$ for each class separately. Finally, the top-100 scoring predictions are selected for each image. For mask inference, we use the top 50 bounding box predictions as mask proposals. Although there are more intelligent ways to perform post-processing, such as SoftNMS~\cite{bodla2017softnms} or test-time image augmentations, in order to fairly compare against the baseline models in speed and accuracy, we intentionally do not use those here.


\vspace{-.1cm}
\section{Experiments}
\label{sec:experiment}
\vspace{-.1cm}

\noindent {\bf Dataset:}
In this paper, we use the COCO~\cite{lin2014coco} dataset, which provides bounding box and segmentation mask annotations. We follow common practice~\cite{bell16ion}, using the COCO \texttt{trainval135k} split (union of 2014train 80k and a subset of 35k images from 2014val 40k) for training and the \texttt{minival} (remaining 5k images from 2014val 40k) for evaluation\footnote{COCO \texttt{trainval135k} is also called COCO 2017 train and the \texttt{minival} is  COCO 2017 val.}.  

\vspace{-.1cm}
\subsection{Ablation Study}
\vspace{-.1cm}

\begin{table}
    \centering
    \begin{tabular}{l|l|l|l|l|l|l}
        Threshold & AP & AP$_{\text{50}}$ & AP$_{\text{75}}$ & AP$_{\text{S}}$ & AP$_{\text{M}}$ & AP$_{\text{L}}$  \\
        \hline
        0.5 & 35.5 & 53.7 &  38.1&19.5&39.3&47.4\\ 
        \hline
        0.4 & 36.0 & 54.1 & 38.6 & 19.1 & 39.6 & 48.3 \\
        \hline
        0.3 & 36.1 & 54.5 & \textbf{38.9} & 19.8 & 39.6 & \textbf{48.6} \\
        \hline
        0.2 & 36.1 & 54.5 & 38.7 & \textbf{20.4} & \textbf{39.8} & \textbf{48.6} \\ 
        \hline
        0.0 & \textbf{36.2} & \textbf{55.0} & 38.7 & 19.7& 39.5 & \textbf{48.6} \\
    \end{tabular}
    \caption{Ablation study of different thresholds used in the best matching case on COCO \texttt{minival}. The selected threshold relaxes the regular intersection-over-union threshold of 0.5 for assigning at least one anchor box to each ground truth box. The base threshold is kept at 0.5, so the modification only affects previously unmatched ground truth objects.}
    \label{tab:bestmatching}
    \vspace{-2.5mm}
\end{table}

\noindent {\bf Best Matching Policy:}
In Table~\ref{tab:bestmatching}, we test the effectiveness of using Best Matching Policy for all ground truth objects, as described in Section~\ref{sec:best_matching}. Threshold $0.5$ corresponds to regular matching. We then gradually lower the threshold for the best matching policy, going down all the way to 0 (completely relaxing the threshold). According to Table~\ref{tab:bestmatching}, using best matching anchors with any positive overlap to ground truth gives the best performance. 

Qualitative results suggest that our matching strategy reduces the number of duplicate detections. Indeed, best matching enforces larger changes to anchor boxes (but not too large to destabilize the training process), so different anchors shrink tighter to the ground truth object, and only one survives during non-maximum suppression.

\smallskip
\smallskip
\noindent {\bf Self-Adjusting Smooth L1 Loss:}
We first ran the Smooth~L1 with fixed values (1.0 and 0.11). 
The choice of $\beta$ is not specificed in RetinaNet~\cite{lin2017focal}. According to the released implementation, Detectron, $0.11$ is used. 
The upper part of Table~\ref{tab:adjustl1} shows that setting $\beta$ is set to $1.0$ will favor will favor $AP_{0.5}$ which is widely used in datasets which adopt IOU=0.5 as the evaluation metric such as Pascal VOC~\cite{Everingham10}. In contrast, the smaller value, $0.11$, will favor more restrictive metrics such as   IOU=.50:.05:.95~\cite{lin2014coco}. 
 
The bottom part of Table~\ref{tab:adjustl1} shows the results of  using Self-Adjusting Smooth L1 loss.
First, we can see that our Self-Adjusting loss with setting 0.11, gives the best results for every metric. It is clear that this method is not dataset dependent.  Second, the Self-Adjusting Smooth L1 is robust. When we change the bounding region from 0.11 to 1.0, the
decrease of results is minor compared to the original Smooth L1 method. We also tried to share the running mean and variance across channels. The result (36.4~mAP) is slightly worse than the separate channel version. 
\begin{table}
    \centering
    \setlength\tabcolsep{5.0pt}
    \begin{tabular}{l|l|l|l|l|l|l}
         & AP & AP$_{\text{50}}$ & AP$_{\text{75}}$ & AP$_{\text{S}}$ & AP$_{\text{M}}$ & AP$_{\text{L}}$ \\
         \hline
         \textit{Fixed} & & & & & &  \\
         $\beta=1.0$ & 35.3 & 55.6 & 37.8 & 19.4 & 38.9 & 46.9 \\
         $\beta=0.11$ &  36.2 & 55.0 & 38.7 & 19.7& 39.5 &48.6 \\
        \hline
        \textit{Self-Adj} & & & & & & \\
        $\beta \leq 1.0$   & 36.4 & 55.4 & \textbf{39.0} & 19.9 & 39.9 & 48.1 \\
        $\beta \leq 0.11$ &  \textbf{36.6} & \textbf{55.7} & \textbf{39.0} & \textbf{20.3} & \textbf{40.0} & \textbf{48.8} \\

    \end{tabular}
    \caption{Ablation study of Self-Adjusting Smooth L1 with different $\beta$ on COCO \texttt{minival}.}
    \label{tab:adjustl1}
    \vspace{-2.5mm}
\end{table}

\begin{table}
\small
\setlength\tabcolsep{5.0pt}
\begin{tabular}{l|l|c|c|c|c|c|c}
Method & Train & AP$^{\text{bb}}$ & AP$^{\text{bb}}_{\text{50}}$ & AP$^{\text{bb}}_{\text{75}}$   &
AP$^{\text{m}}$ & 
AP$^{\text{m}}_{\text{50}}$ & 
AP$^{\text{m}}_{\text{75}}$ \\
\hline
Base & 1x & 36.2 & 55.0 & 38.7 & \textemdash & \textemdash & \textemdash \\
P$_3$-P$_5$& 1x & 36.9&55.3& \textbf{39.7} & 32.7&52.2&34.9 \\
P$_3$-P$_5$ & 1.5x & \textbf{37.1} & \textbf{55.9} & 39.5 & \textbf{33.0} & \textbf{52.9} & 35.0 \\
P$_2$-P$_5$& 1x &  36.7 & 54.9 & \textbf{39.7}& 32.8&52.2& \textbf{35.1}\\

\end{tabular}
\caption{Ablation study of different settings for adding mask prediction module on COCO \texttt{minival}.}
\label{table:mask_training}
\vspace{-2.5mm}
\end{table}

\smallskip
\smallskip
\noindent {\bf Multi-Task Training with Mask Prediction:}
Table~\ref{table:mask_training} illustrates bounding box accuracy improvement when running multi-task training with instance segmentation. When training with mask prediction using \{P$_{3}$, P$_{4}$, P$_{5}$\}, we see 0.7~mAP improvement. If we train with 1.5x schedule,  the improvement is 0.9~mAP. If we add the feature layer with higher resolution, will it be helpful for the prediction? We follow Mask R-CNN to use \{P$_{2}$, P$_{3}$, P$_{4}$, P$_{5}$\} for mask prediction. The results are slightly better on mask prediction but worse on detection. 
In conclusion, adding mask prediction consistently improves detection results, but requires longer training. It is also worth noting that in the Mask R-CNN ablation study, the authors also show 0.9~mAP improvement on bbox prediction from multi-tasking training.

\begin{figure}[h]
    \begin{subfigure}{\linewidth}
        \begin{subfigure}{0.48\textwidth} 
            \centering
            \includegraphics[width=\textwidth, height=2.6cm]{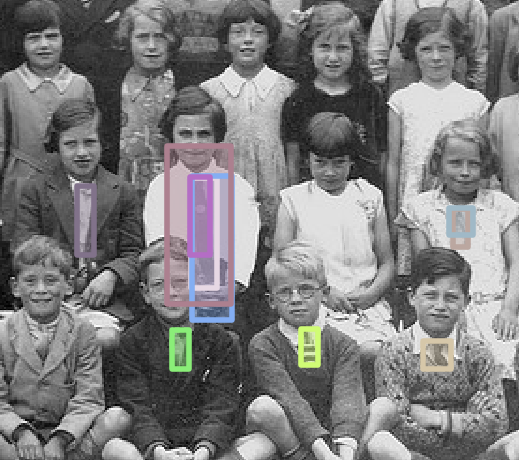}
        
        \end{subfigure}%
        \hfill
        \begin{subfigure}{0.48\textwidth}
            \centering
            \includegraphics[width=\textwidth, height=2.6cm]{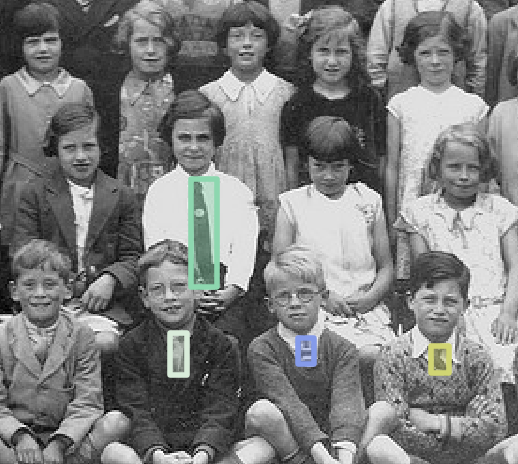}

        \end{subfigure}
    \caption{Tie}
    \end{subfigure}
    \begin{subfigure}{\linewidth}
        \begin{subfigure}{0.48\textwidth}
            \centering
            \includegraphics[width=\textwidth, height=2.6cm]{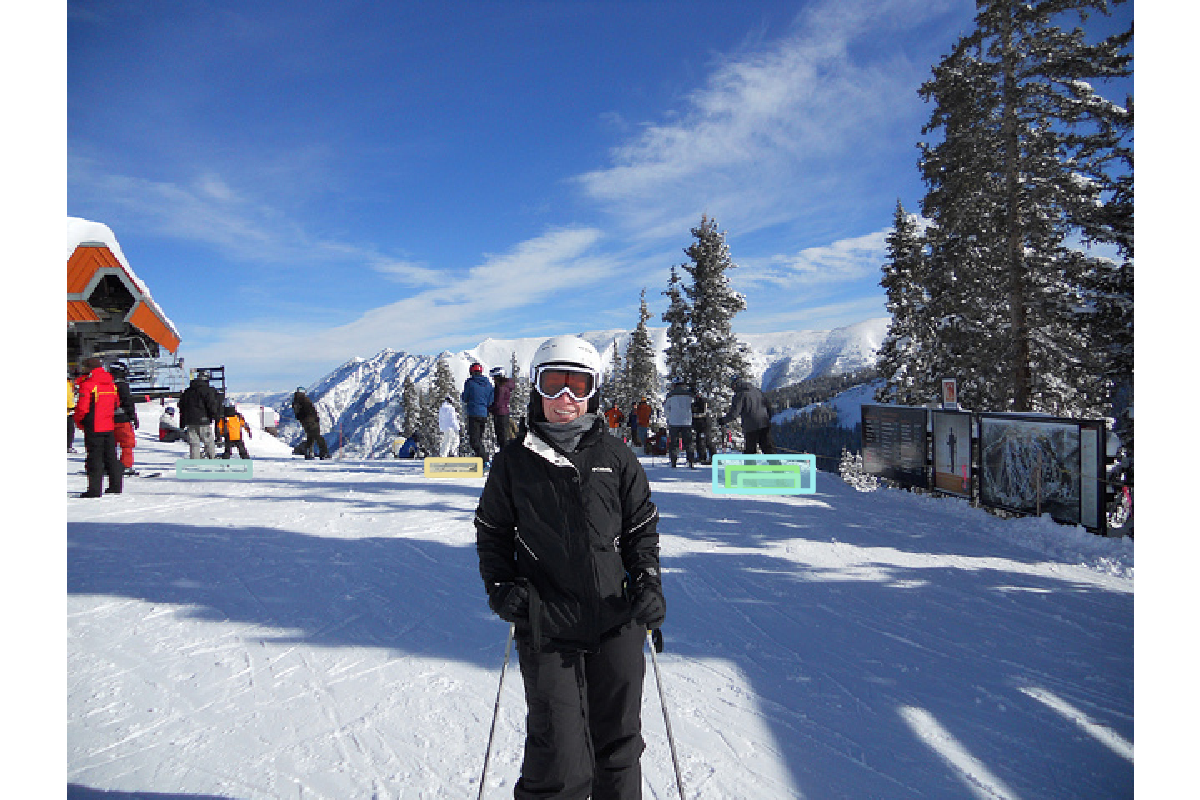}
        \end{subfigure}%
        \hfill
        \begin{subfigure}{0.48\textwidth}
            \centering
            \includegraphics[width=\textwidth, height=2.6cm]{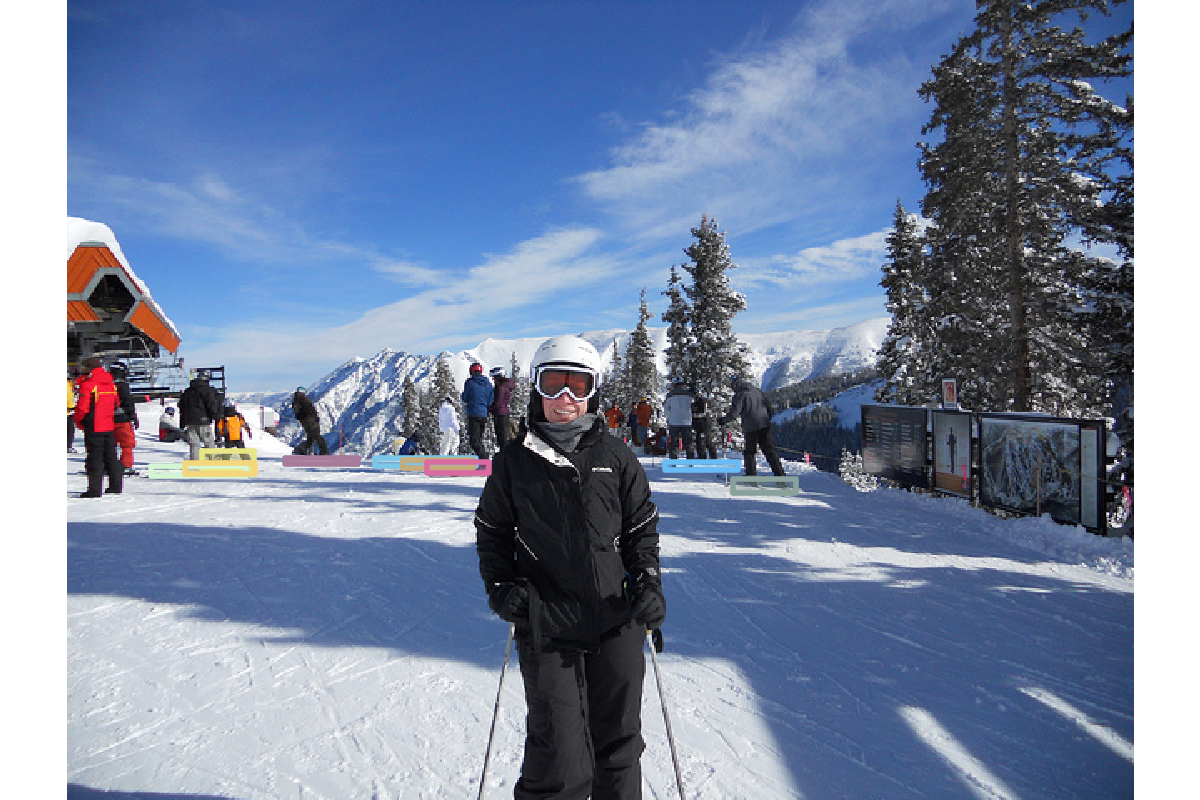}
        \end{subfigure}
        \caption{Ski}
    \end{subfigure}
    \begin{subfigure}{\linewidth}
        \begin{subfigure}{0.48\textwidth}
            \centering
            \includegraphics[width=\textwidth, height=2.6cm]{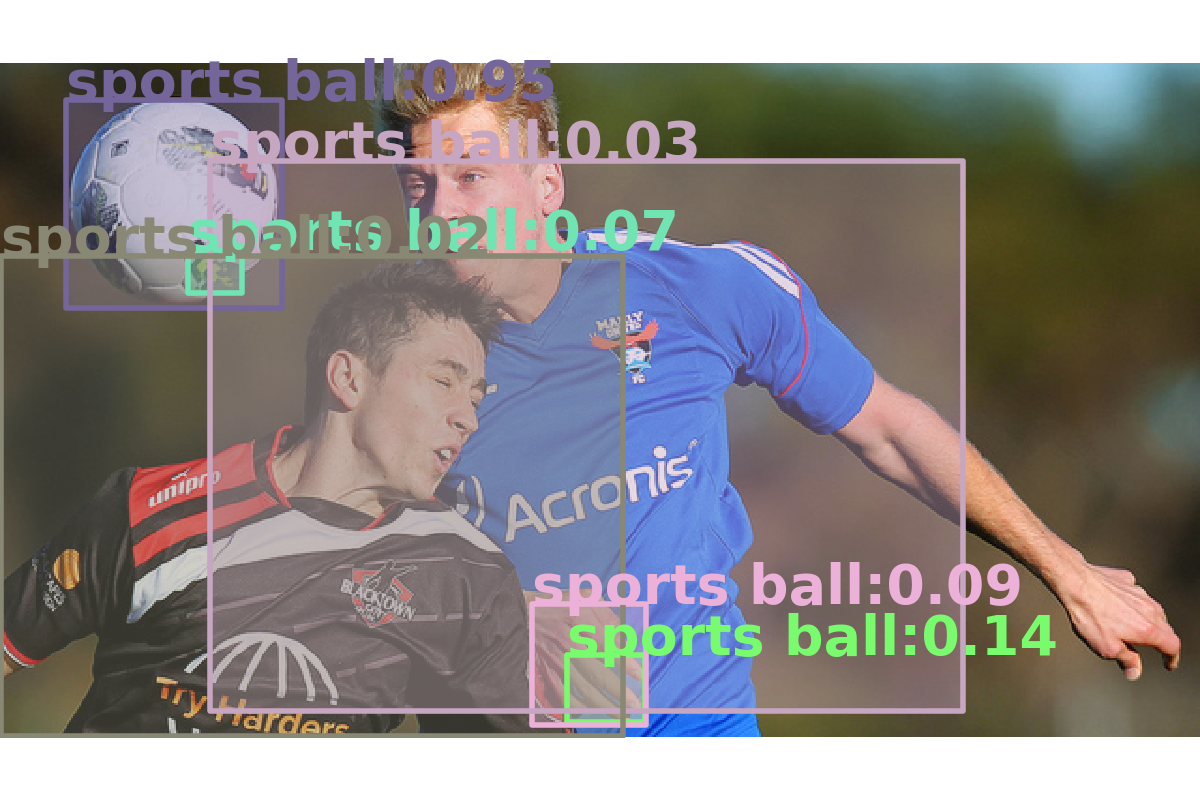}
        \end{subfigure}%
        \hfill
        \begin{subfigure}{0.48\textwidth}
            \centering
            \includegraphics[width=\textwidth, height=2.6cm]{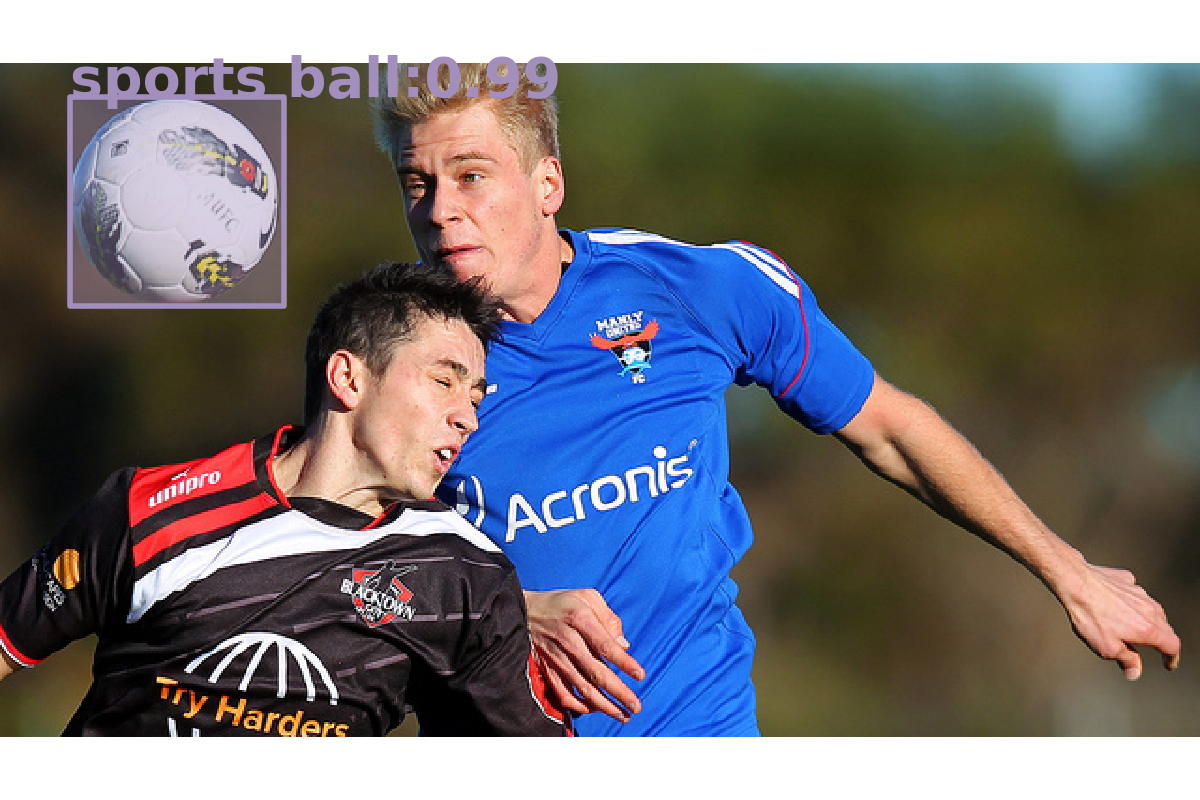}
        \end{subfigure}
        \caption{Sports Ball}
    \end{subfigure}
    \begin{subfigure}{\linewidth}
        \begin{subfigure}{0.48\textwidth}
            \centering
            \includegraphics[width=\textwidth, height=2.6cm]{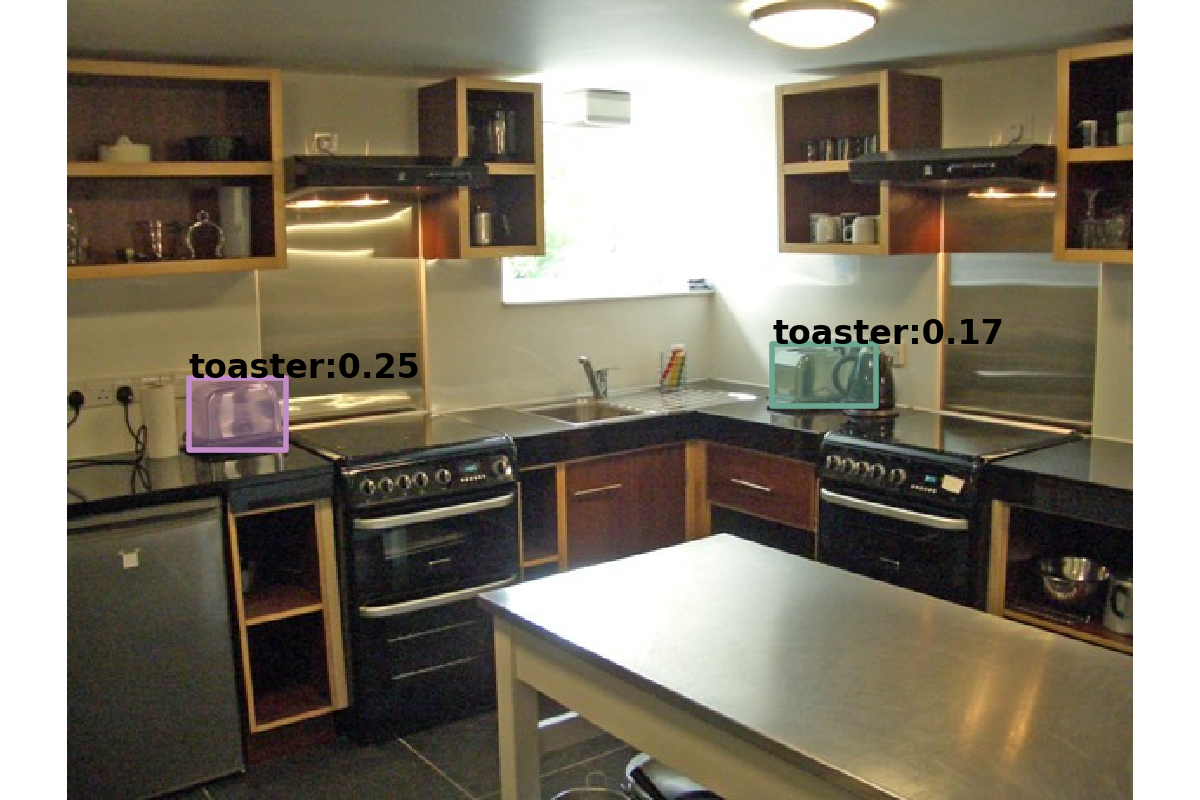}
        \end{subfigure}%
        \hfill
        \begin{subfigure}{0.48\textwidth}
            \centering
            \includegraphics[width=\textwidth, height=2.6cm]{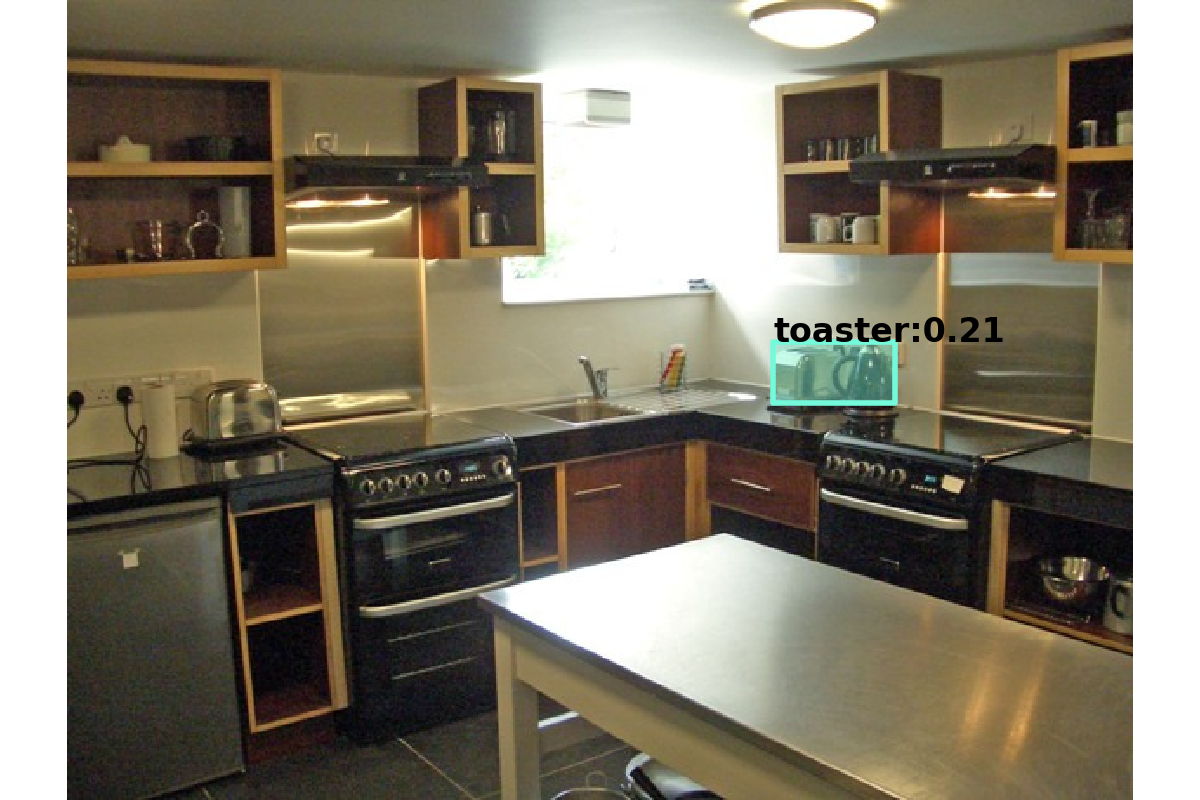}
        \end{subfigure}
        \caption{Toaster}
    \end{subfigure}
    \caption{Comparison to RetinaNet baseline (left column). RetinaMask (right column) includes Best Matching policy, Self-Adjusting L1 loss, and Mask Prediction (see Section~\ref{sec:compare_with_retinanet}). The figure shows all detection results (no confidence threshold applied) only for selected categories (\emph{tie} for (a), \emph{skis} for (b), \emph{sports ball} for (c), and \emph{toaster} for (d)). Prediction scores are also shows, where they do not clutter the image. Our model shows improvement in classes with large aspect ratios (no multiple detections for \emph{tie}, and better recall for \emph{skis}); In (a) our model demonstrates no false negatives (e.g. note false negative sports ball with $0.14$ score for the baseline model); (d) shows the failure toaster case, that accounts for the decrease in Figure~\ref{fig:mAP_all} (only 9 toasters in COCO \texttt{minival}). Better viewed electronically, enlarged.}
    \label{fig:comparison_examples}
    \vspace{-2.5mm}
\end{figure}

\vspace{-.1cm}
\subsection{Comparison to RetinaNet}
\label{sec:compare_with_retinanet}
\vspace{-.1cm}

Following ~\cite{hoiem2012diagnosing} we give an explanation of our model's improvement over the RetinaNet baseline. The model evaluated in this section incorporates all three components described in Section~\ref{sec:model}: Best Matching policy, Self-Adjusting Smooth L1, and Mask Prediction head. ResNet-50 is used as the backbone architecture, and images are resized to a shorter side of 800 pixels. No data augmentation is used.

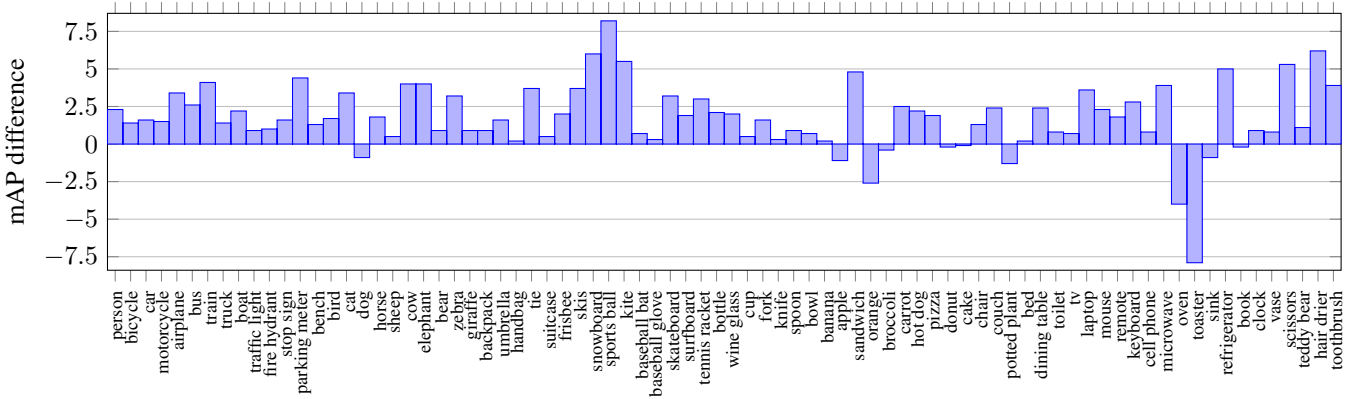
\begin{figure*}
\begin{tikzpicture}
\begin{axis}[
ybar, 
bar width=0.205cm, 
height=5cm,
x=0.205cm,
ylabel={mAP difference},
flexible xticklabels from table={tables/mAP_all.csv}{name}{col sep=comma},
xticklabel style={text height=1.5ex, rotate=90,anchor=east, 
font=\scriptsize,}, 
ymajorgrids=true,
ytick={-7.5, -5, -2.5,0,2.5, 5, 7.5, 10},
xtick=data,
enlarge x limits={abs=0.5},
enlarge y limits={abs=0.5},
]
\addplot table[x expr=\coordindex,y=y,]{\datatable} ;

\end{axis}
\end{tikzpicture}
\caption{RetinaMask mAP detection improvement over RetinaNet baseline ResNet-50 backbone results are shown. mAP is computed across top 100 detections, and averaged for thresholds in range .50:.05:.95, according to COCO~\cite{lin2014coco}.}
\label{fig:mAP_all}
\end{figure*}

First, we look at per-class difference of the mean Average Precision  in Figure~\ref{fig:mAP_all}, showing improvement in most of the classes. Note that the \emph{toaster} class, whose mAP decreases by $7.9$ points (from 28.9 to 21.0), has only $9$ ground truth objects in the validation set. On the other hand, hair drier shows a significant increase from $0.9$ to $7.1$ mAP points. The classes that improve most also include \emph{snowboard}, \emph{sports ball}, \emph{kite}, \emph{refrigerator}, and \emph{scissors} (mAP difference $\geq$ 5). See some qualitative examples in Figure~\ref{fig:comparison_examples}.

\begin{figure}[h!]
    \begin{subfigure}{0.5\textwidth}
        \includegraphics[width=0.49\textwidth]{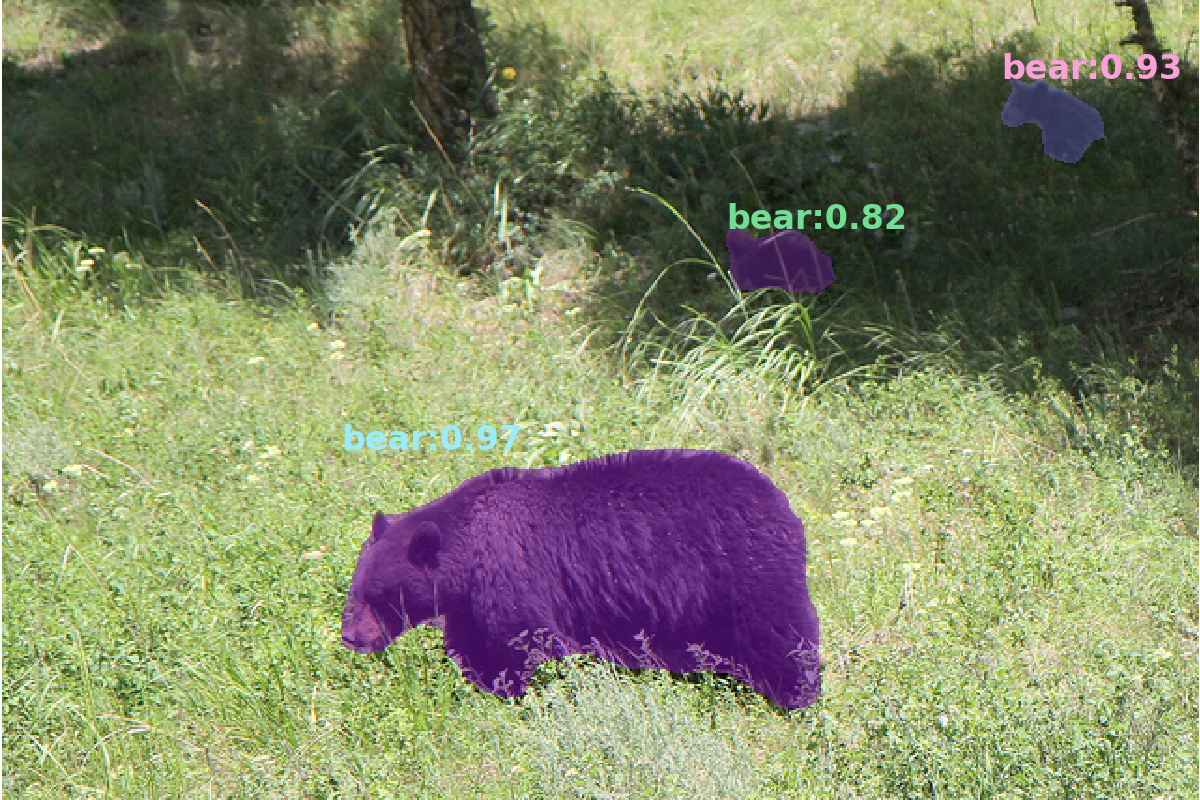}
        \includegraphics[width=0.49\textwidth]{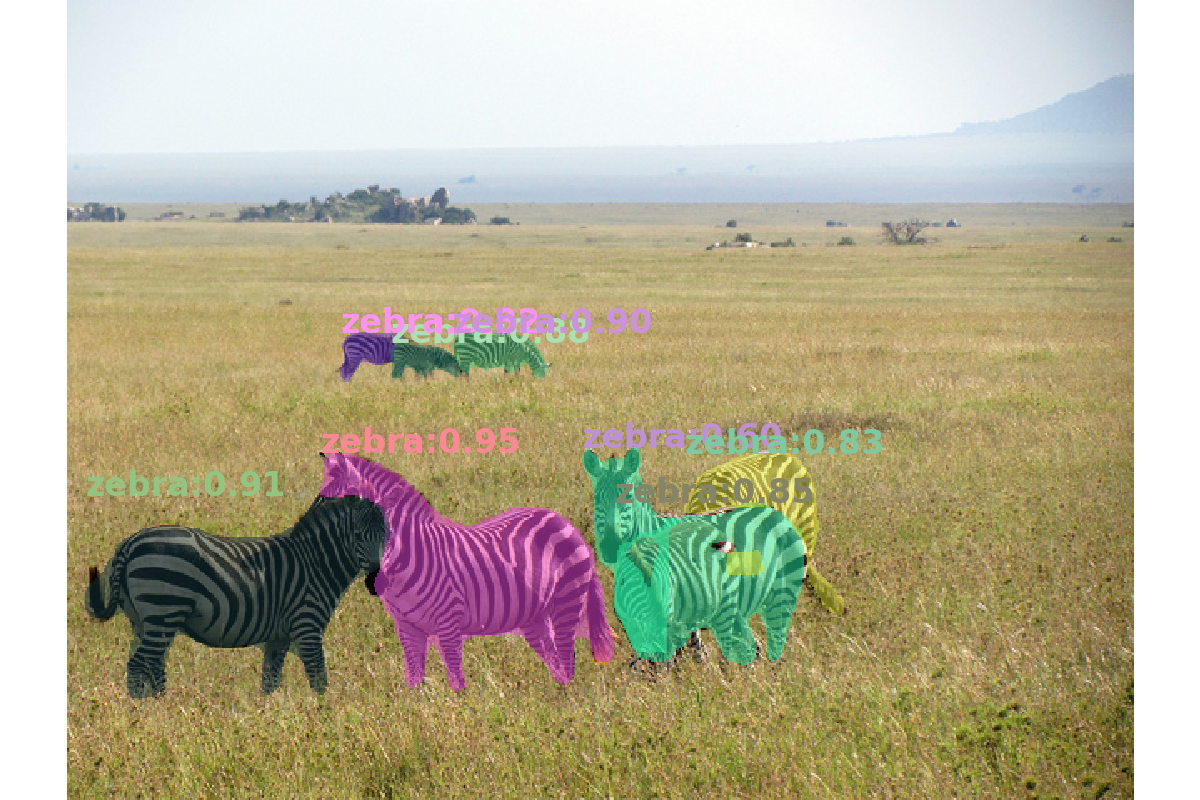}
    \end{subfigure}
    \begin{subfigure}{0.5\textwidth}
        \includegraphics[width=0.49\textwidth]{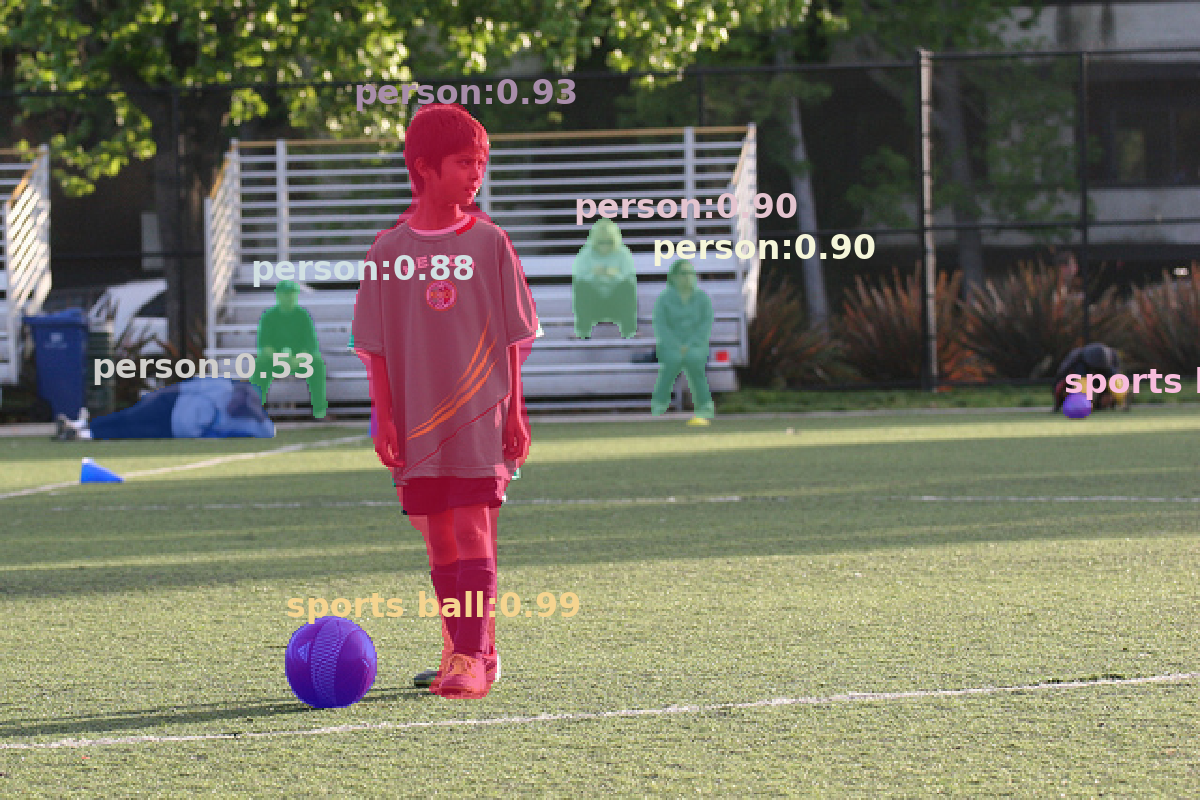}
        \includegraphics[width=0.49\textwidth]{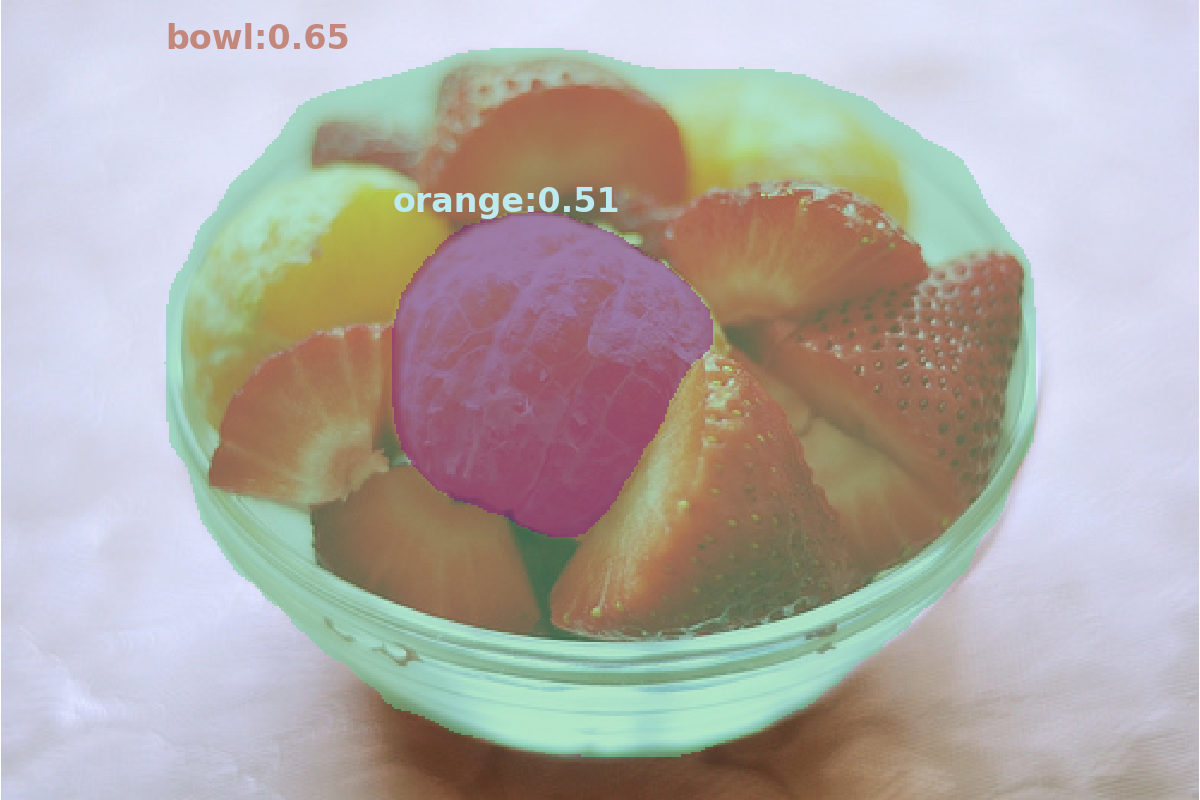}
    \end{subfigure}
    
    \begin{subfigure}{0.5\textwidth}
         \includegraphics[width=0.49\textwidth]{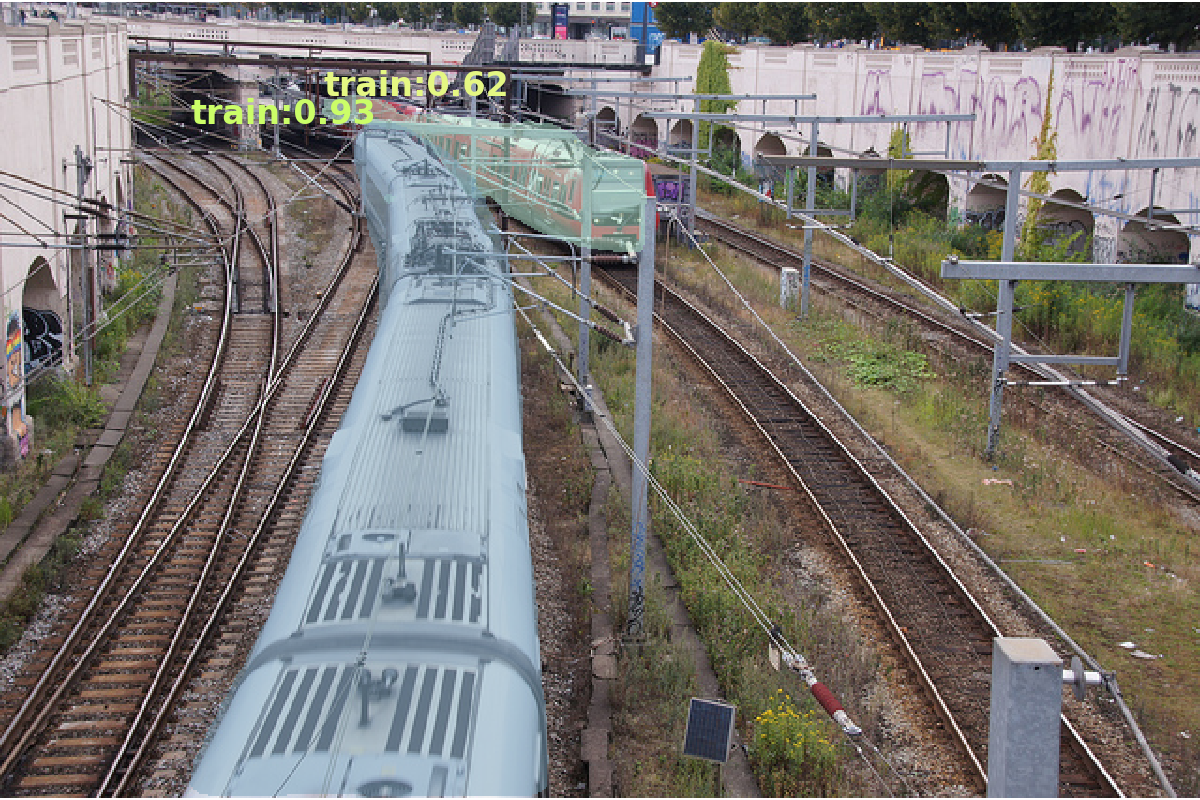}
          \includegraphics[width=0.49\textwidth]{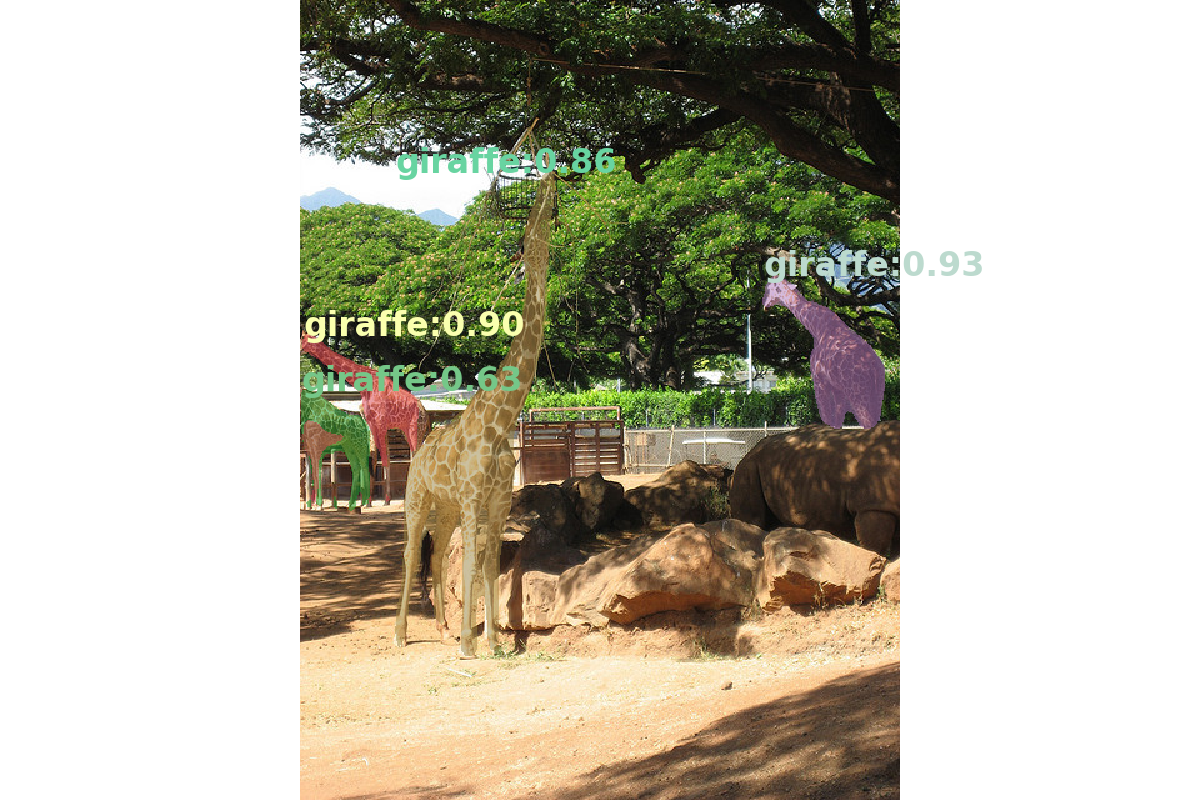}
    \end{subfigure}
    \caption{Visualization of RetinaMask with ResNet-101-FPN-GN Model(BBox=41.7~mAP, Mask=36.7~mAP result on COCO \texttt{test-dev}). }
    \label{fig:final_examples}
    \vspace{-2.5mm}
\end{figure}

\begin{figure*}[h]

\begin{tikzpicture}
\begin{axis}[
ybar, 
bar width=0.205cm, 
x=0.205cm,
ylabel={mAP difference},
flexible xticklabels from table={tables/oth_all.csv}{name}{col sep=comma},
xticklabel style={text height=1.5ex, rotate=90,anchor=east, 
font=\scriptsize,}, 
ymajorgrids=true,
ytick={-7.5, -5, -2.5,0,2.5, 5, 7.5, 10, 12.5, 15, 17.5},
xtick=data,
enlarge x limits={abs=0.5},
enlarge y limits={abs=0.5},
]
\addplot table[x expr=\coordindex,y=y,]{\datatableoth} ;

\end{axis}
\end{tikzpicture}
\caption{RetinaMask class-agnostic detection improvement over RetinaNet baseline. Localization errors are ignored by setting a low IoU threshold of 0.1, foreground object mis-classification is ignored as well. ResNet-50 backbone results are shown.}
\label{fig:oth_all}
\end{figure*}
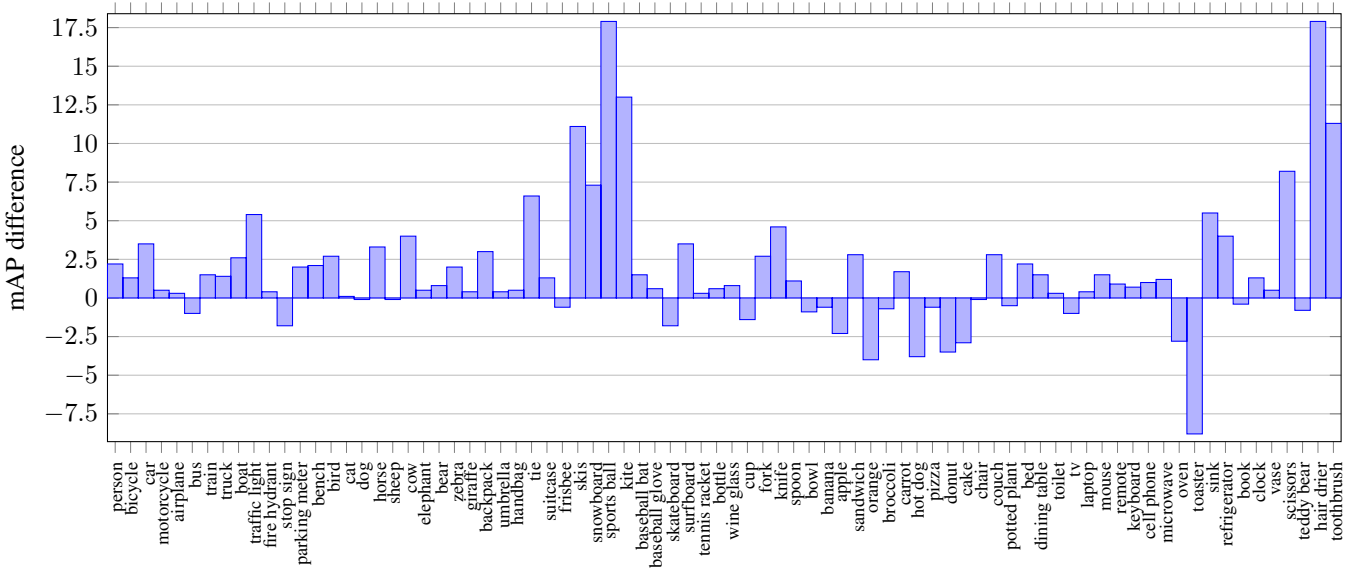

Figure~\ref{fig:oth_all} shows the difference in class-agnostic weak detection at low IoU threshold of 0.1, which ignores localization errors. Moreover, foreground object mis-classification is also ignored, which does not count for errors of attributing an object of one category to a different category. High correlation in these two relative differences for the improved classes (\emph{snowboard}, \emph{sports ball}, etc.) suggest that a large portion of network improvement comes from better localization, rather than better confidence prediction (otherwise class-agnostic weak detection would not improve).

\begin{table}[h]
\footnotesize
    \centering
    \setlength\tabcolsep{4.5pt}
    \begin{tabular}{c|c|c|c|c|c|c|c|c|r }
         D & S & M & AP & AP$_{\text{50}}$ & AP$_{\text{75}}$ & AP$_{\text{S}}$ & AP$_{\text{M}}$ & AP$_{\text{L}}$ & T \\
         \specialrule{.2em}{.1em}{.1em} 
         50 & 400 & R & 30.5 & 47.8 & 32.7 & 11.2 & 33.8 & 46.1 & 64\\
         && O(B) & 32.3 & 49.8 &34.2 & 11.6 & 34.4 & 47.8 & 73\\
         && O(M)  & 28.7 & 46.9 & 30.1 & 07.0 & 29.1 & 47.4  & 82\\ 
         \hline
         
         50 & 500 & R & 32.5 & 50.9 & 34.8 & 13.9 & 35.8 & 46.7 & 72 \\
         && O(B) & 34.6 & 52.6 & 36.7 & 14.8 & 36.7 & 48.8& 81\\
         && O(M) & 30.7 & 49.8 & 32.3 & 09.6 & 31.7 & 48.2 & 92\\
        
         \hline
         50 & 600 & R & 34.3 & 53.2 & 36.9 & 16.2 & 37.4 & 47.4& 98\\
         &  & O(B) & 36.0 & 54.5 & 38.5 & 17.1 & 38.1 & 49.2 & 97\\
         & &  O(M) & 31.9 & 51.6 & 33.8 & 11.5 & 33.3 & 48.4 & 108\\
         \hline 
        50 & 700 & R & 35.1 & 54.2 & 37.7 & 18.0 & 39.3 & 46.4 & 121 \\
        && O(B) & 36.9 & 55.6 & 39.6 & 18.9 & 39.1 & 48.7 &  111\\
        &&  O(M) &   32.8 & 52.9 & 35.0 & 13.0 & 34.5 & 48.4 & 123\\
        
        \hline
         
         50 & 800 & R & 35.7 & 55.0 &  38.5 & 18.9 & 38.9 & 46.3 & 153\\
         & & O(B) & 37.5  & 56.4 & 40.2 & 19.6 & 39.8 & 48.9 &   124\\
         & &  O(M)  & 33.4 & 53.7 & 35.4 & 13.6 & 35.1 & 48.7 & 141 \\
         \hline
         \hline
         
        101 & 400 & R & 31.9 & 49.5 & 34.1 & 11.6 & 35.8 & 48.5 & 81\\
         && O(B) & 33.1 & 49.7 & 35.4 & 11.0 & 35.7 & 49.9 & 87	 \\
         && O(M) & 29.4 & 47.3 & 31.2 & 06.8 & 30.4 & 49.5 & 99\\ 
         \hline
         
        101 & 500 & R & 34.4 & 53.1 & 36.8 & 14.7 & 38.5 & 49.1 & 90\\
         && O(B) & 36.3 & 54.7 & 38.7 & 15.9 & 39.1 & 50.9 & 93\\
         && O(M) & 32.0 & 51.8 & 33.8 & 10.2 & 33.8 & 50.2 & 105\\ 
        \hline
         
         101 & 600 &  R & 36.0 & 55.2 & 38.7 & 17.4 & 39.6 & 49.7 & 122\\
         &  & O(B) &37.4  & 56.0 & 39.9 & 17.3 &39.9 & 51.4  & 110\\
         & & O(M) & 33.2 & 53.2 & 35.2 & 11.6 & 35.0 & 50.7 & 120\\
         \hline
         
        101 & 700 & R & 37.1 & 56.6 & 39.8 & 19.1 & 40.6 & 49.4 & 154\\ 
        && O(B) & 38.5 & 57.3 & 41.3 & 19.1 & 41.1 & 51.8 & 126\\
        && O(M) & 34.1 & 54.5 & 36.3 & 13.0 & 36.2 & 51.0 & 137\\ 
        \hline

         101 & 800 &  R & 37.8 & 57.5 & 40.8 & 20.2 & 41.1 & 49.2 & 198\\
        && O(B) & 39.1 & 58.0 & 41.9 & 20.4 & 41.7 & 51.0 & 145\\
        & &  O(M) & 34.7 & 55.4 & 36.9 & 14.3 & 36.7 & 50.5 & 166 \\

    \end{tabular}
    \caption{Comparison to RetinaNet with different input resolutions on COCO \texttt{test-dev} (Also see Figure~\ref{fig:speed_vs_accuracy}). For each (D/S) depth/scale, the upper part (R) is the RetinaNet performance from Table 1(e) in RetinaNet~\cite{lin2017focal}, our results are in the bottom part. We also report mask prediction accuracies. For Depth, 50:ResNet-50-FPN, 101:ResNet-101-FPN. For Method, R:RetinaNet, O(B): Our Method of BBox prediction,  O(M): Our Method of Mask prediction. Our speed number is evaluated on Nvidia 1080 Ti / PyTorch1.0 and FocalLoss results are evaluated on Nvidia M40 / Caffe2.}
    \label{tab:compare_diff_resolutions}
\end{table}

\begin{table*}
\centering
\footnotesize
\begin{tabular}{l|l|lll|lll}
Method & Backbone & AP$^{bb}$ &
AP$^{bb}_{\text{50}}$ & 
AP$^{bb}_{\text{75}}$ &
AP$^{S}_{\text{75}}$& 
AP$^{M}_{\text{75}}$& 
AP$^{L}_{\text{75}}$ \\
\hline 
\textit{Two-Stage Detectors} & & & & & & & \\
Faster R-CNN+++~\cite{he2015resnet} & ResNet-101-C4 & 34.9 & 55.7 & 37.4 & 15.6 & 38.7 & 50.9 \\
Faster R-CNN w FPN~\cite{lin2016fpn} &  ResNet-101-FPN & 36.2 & 59.1 & 39.0 & 18.2 & 39.0 & 48.2 \\

Faster R-CNN w RoIAlign~\cite{he2017maskrcnn} & ResNet-101-FPN & 37.3 & 59.6 & 40.3 & 19.8 & 40.2 & 48.8 \\
Mask R-CNN~\cite{he2017maskrcnn} & ResNet-101-FPN & 38.2 & 60.3 & 41.7 & 20.1 & 41.1 & 50.2\\

\hline
\textit{single-shot Detectors}  &  & & & & & & \\
YOLOv2~\cite{redmon2017yolov2} & Darknet-19 & 21.6 & 44.0 & 19.2 & 5.0 & 22.4  & 35.5 \\
SSD513~\cite{liu2016ssd,fu2017dssd}  & ResNet-101 & 31.2 & 50.4 & 33.3 & 10.2 & 34.5 & 49.8 \\
DSSD513~\cite{fu2017dssd} & ResNet-101-DSSD & 33.2 & 53.3 & 35.2 & 13.0 & 35.4 & 51.1\\
YOLOv3-608~\cite{redmon2018yolov3} & Darknet-53 & 33.0 & 57.9 & 34.4 & 18.3 & 35.4 & 41.9 \\
RetinaNet~\cite{lin2017focal} & ResNet-101-FPN & 39.1 & 59.1 & 42.3 & 21.8 & 42.7 & 50.2 \\
RetinaNet~\cite{lin2017focal} & ResNeXt-101-FPN & 40.8 & 61.1 & 44.1 & 24.1 & 44.2 &  51.2 \\
RetinaMask & ResNet-50-FPN & 39.4 & 58.6 & 42.3 & 21.9 & 42.0 & 51.0 \\
RetinaMask & ResNet-101-FPN & 41.4 & 60.8 & 44.6 & 23.0 & 44.5 & 53.5 \\
RetinaMask & ResNet-101-FPN-GN & 41.7 & 61.7 & 45.0 & 23.5 & 44.7 & 52.8 \\
RetinaMask & ResNeXt-101-FPN-GN & \textbf{42.6} & \textbf{62.5} & \textbf{46.0} & \textbf{24.8} & \textbf{45.6} & \textbf{53.8}\\

\end{tabular}
\caption{Comparison with state-of-the-art methods on COCO \texttt{test-dev}. Compared to RetinaNet~\cite{lin2017focal}, our model based on ResNet-101-FPN is better by 2.6~mAP. Compared to Mask R-CNN~\cite{he2017maskrcnn}, our model shows 3.5~mAP improvement. }
\label{table:compare_stoa}
\vspace{-2.5mm}
\end{table*}

Table~\ref{tab:compare_diff_resolutions} shows  comparisons of our model to RetinaNet on different backbone networks and input resolutions. RetinaNet results come from the Table~1(e) of the RetinaNet~\cite{lin2017focal} paper. Our model shows better accuracy for all combinations of backbone network choices and  resolutions. We report the speed number evaluated on Nvidia 1080~Ti. We re-implement the network in PyTorch (1080Ti), while RetinaNet is implemented in Caffe2~(M40). Note that speed numbers are reported for different GPU architectures, and thus should not be directly compared. Our network is very similar in inference settings to the original RetinaNet, so most speed performance gains are attributed to better framework implementation.

In Figure~\ref{fig:speed_vs_accuracy}, we show our results compared with state-of-the-art single-shot and two-stage detectors~\cite{redmon2018yolov3,lin2017focal,he2017maskrcnn}. Note that YOLOv3~\cite{redmon2018yolov3} is trained with multi-scale training but ours and ReinaNet~\cite{lin2017focal} are not. Our results show that the detector in RetinaMask  has a higher envelop for accuracy-vs-time than RetinaNet when using ResNet-50 and ResNet-101 for the backbone model. All the numbers for Figure~\ref{fig:speed_vs_accuracy} can be found in Table~\ref{tab:compare_diff_resolutions}. Our model shows 1.84~mAP and 1.52~mAP improvement on ResNet-50 and ResNet-101 compared to RetinaNet. Our detection result is better than the original numbers from Mask R-CNN and very close to recent implementation results. 

Note that the speed of our implementation on the short side of 400 is surprisingly slow.  We think this is an idiosyncrasy of the libraries used, and note that as with all the other resolutions we do see an improvement in accuracy.



%

\vspace{-.1cm}
\subsection{Comparisons to the state-of-the-art methods}
\vspace{-.1cm}
We use ResNet-50-FPN, ResNet-101-FPN, and ResNeXt32x8d-101-FPN~\cite{Xie2016resnext} as the backbones in our final models. We train with the multi-scale \{640, 800, 1200\} and 2x iterations schedule. For the ResNet-101-FPN model, we also train a version using Group Normalization(GN)~\cite{wu2018gn}, which is applied only on the extra layers (FPN, localization, and classification). 
Replacing all the Batch Normalization~\cite{ioffe2015bn} in ResNet-101 would cause a significant slowdown. The speed of ResNet-101-FPN-GN model is 0.158~s/im (compared to 0.145~s/im without GN). Using ResNeXt32x8d-101-FPN~\cite{Xie2016resnext} as  backbone further improves results by 0.9~mAP and achieves 42.6~ mAP on COCO.  We provide the quantitative comparison in Table~\ref{table:compare_stoa} and show some detection examples in Figure~\ref{fig:final_examples}.

We also acknowledge the recent new architectures for better object detection such as NASNET~\cite{zoph2018nasnet} or efficient networks (MobileNet~\cite{howard2017mobilenet,sandler2018mobilenetv2}, ShuffleNet~\cite{zhang2018shufflenet}), but their evaluation is beyond the scope of this work.

\vspace{-.1cm}
\subsection{Comparison with Mask R-CNN on instance mask prediction}
\vspace{-.1cm}
In Table~\ref{tab:Compare with Mask R-CNN}, we compare our mask (instance segmentation) results to Mask R-CNN. The Mask R-CNN~\cite{he2017maskrcnn} results are from Table~8 of Mask R-CNN~\cite{he2017maskrcnn}. All the results are using ResNet-101 and Feature Pyramid Network~\cite{lin2016fpn} as the backbone model. Our models are trained in a very similar fashion to the \textit{+e2e training} in ~\cite{he2017maskrcnn}. Mask R-CNN still shows better accuracy on mask prediction, but the difference is only around 1.2~mAP.

\begin{table}
\small
\begin{tabular}{l|l|l|l|l|l|l}
Method & AP$^{\text{m}}$ & AP$^{\text{m}}_{\text{50}}$ & AP$^{\text{m}}_{\text{75}}$   &
AP$^{\text{bb}}$ & 
AP$^{\text{bb}}_{\text{50}}$ & 
AP$^{\text{bb}}_{\text{75}}$ \\
\hline
\textit{Mask R-CNN} &36.7 &59.5 &38.9 &39.6 &61.5 &43.2 \\
+update baseline & 37.0&59.7&39.0&40.5&63.0&43.7 \\
+e2e training &37.6&60.4&39.9&41.7&64.1 &45.2\\
+ImageNet-5k  & 38.6 & 61.7 & 40.9 & 42.7 & 65.1 & 46.6\\
+train-time augm. & \textbf{39.2} & \textbf{62.5} & \textbf{41.6} & \textbf{43.5} & \textbf{65.9} & \textbf{47.2} \\
\hline
RetinaMask & 36.4 & 57.3 & 38.7 & 41.1 & 60.2 & 44.1\\

\end{tabular}
\caption{Comparison with Mask R-CNN on mask prediction using ResNet-101 on COCO \texttt{minival}. The Mask R-CNN results are from Table~8 in the appendix of Mask R-CNN~\cite{he2017maskrcnn}.}
\label{tab:Compare with Mask R-CNN}
\vspace{-2.5mm}
\end{table}

\vspace{-.1cm}
\section{Conclusion}
\vspace{-.1cm}
In this work, we proposed three components to train a more accurate Single-Shot detector, RetinaMask. Our ablations show improvements for each module and our final model shows better accuracy without any network architecture change during inference. The proposed Self-Adjusting Smooth~L1 loss can be used beyond the tasks of object detection and instance segmentation. 

\vspace{-.2cm}
\section*{Acknowledgements}
\vspace{-.2cm}
We thank Tamara Berg, and Phil Ammirato for their helpful suggestions, and we acknowledge support from NSF 1452851, 1533771, 1526367.

\clearpage

\clearpage

{\small
\bibliographystyle{ieee}
\bibliography{egbib}
}

\end{document}